\documentclass[letterpaper, 10 pt, conference]{ieeeconf}
\IEEEoverridecommandlockouts
\usepackage[space]{cite}
\usepackage[normalem]{ulem}
\usepackage{amsmath,amssymb,amsfonts}
\usepackage{dblfloatfix} 
\usepackage{algorithm}
\usepackage{algorithmic}

\DeclareMathOperator*{\argmin}{argmin}
\DeclareMathOperator*{\argmax}{argmax}
\usepackage{graphicx}
\usepackage{textcomp}
\usepackage{color}
\usepackage{subcaption}
\def\BibTeX{{\rm B\kern-.05em{\sc i\kern-.025em b}\kern-.08em
    T\kern-.1667em\lower.7ex\hbox{E}\kern-.125emX}}
\begin{document}
\title{Human-in-the-Loop Robot Planning with Non-Contextual Bandit Feedback
}

\author{
Yijie Zhou, Yan Zhang, Xusheng Luo, and Michael M. Zavlanos
\thanks{The authors are with the Department of Mechanical Engineering and Material Science, Duke University, Durham, NC 27708, USA {\tt\small \{e-mail: yijie.zhou,yan.zhang2,xusheng.luo,
michael.zavlanos\}@duke.edu}. This work is supported in part by AFOSR under award \#FA9550-19-1-0169
and by NSF under award CNS-1932011.}
}

\maketitle

\begin{abstract}
In this paper, we consider robot navigation problems in environments populated by humans. The goal is to determine collision-free and dynamically feasible trajectories that also maximize human satisfaction, by ensuring that robots are available to assist humans with their work as needed and avoid actions that cause discomfort. In practice, human satisfaction is subjective and hard to describe mathematically. As a result, the planning problem we consider in this paper may lack important contextual information. To address this challenge, we propose a semi-supervised Bayesian Optimization (BO) method to design globally optimal robot trajectories using bandit human feedback, in the form of complaints or satisfaction ratings, that expresses how desirable a trajectory is. Since trajectory planning is typically a high-dimensional optimization problem in the space of waypoints that need to be decided, BO may require prohibitively many queries for human feedback to return a good solution. To this end, we use an autoencoder to reduce the high-dimensional space into a low dimensional latent space, which we update using human feedback. Moreover, we improve the exploration efficiency of BO by biasing the search for new trajectories towards dynamically feasible and collision-free trajectories obtained using off-the-shelf motion planners. We demonstrate the efficiency of our proposed trajectory planning method in a scenario with humans that have diversified and unknown demands. 

\end{abstract}


\section{Introduction}
A growing number of current and future robotics applications, such as unmanned delivery \cite{vyas_unmanned_2016,bamburry_drones_2015} and robotic nurse assistance\cite{gombolay_robotic_2016,mahrouche_robotic_2014,pineau_towards_2003,hu_advanced_2011}, require robots to operate in close proximity to humans. In these applications, trajectory planning does not only need to meet critical safety requirements, such as collision avoidance, but it also needs to ensure that robots are available to assist humans with their work as needed. This may involve adapting to specific ways that specific humans perform their tasks or avoiding actions that cause discomfort. In practice, such requirements are subjective and hard to model mathematically. As a result, new planning methods are needed to design robot trajectories that maximize human satisfaction, using exclusively bandit human feedback that lacks contextual information typically captured by mathematical models.

Existing motion planning methods in environments populated by humans usually model humans as moving obstacles and, therefore, reduce the planning problem to a standard collision-avoidance problem. 
For example, the methods in \cite{chen2017socially,r33} use social force models or non-parametric models to predict the movements of humans and then plan robot trajectories using the learnt human models. A different approach is proposed in \cite{chen_socially_2018} that relies on reinforcement learning (RL) to train a navigation policy using deep neural networks. Similarly, the method proposed in \cite{kim_socially_2016} learns to control the robot in a human-populated environment from human demonstrations using inverse RL. Nevertheless, by modeling humans as moving obstacles and solving collision avoidance problems, it is not possible to  satisfy more general human needs. Since human satisfaction is subjective and hard to model mathematically, a much more effective approach is to formulate planning problems that use bandit human feedback in the form of complaints or satisfaction rankings. 

Robot planning using bandit human feedback has been studied in \cite{daniel_active_2015,menon_towards_2015,luo_socially-aware_2020}. 
Specifically, in \cite{daniel_active_2015}, a robot simultaneously learns a policy and a reward model by actively querying a human expert for ratings. Similarly, \cite{menon_towards_2015} adopts a data-driven approach to design robot trajectories using human ratings on different roll outs. Nevertheless, these learning-based methods often require prohibitively many queries for human feedback, especially for high dimensional problems.
In our recent work \cite{luo_socially-aware_2020} we proposed a new zeroth-order optimization method that uses bandit human feedback to design robot trajectories that increase human satisfaction. However, since this planning problem is highly non-linear, zeroth-order methods can get trapped at local minima, reducing the quality of the resulting paths. Moreover, the method in \cite{luo_socially-aware_2020} only considers a single type of human feedback, specifically human complaints when the robot interferes with human tasks.

To plan globally optimal robot trajectories that maximize human satisfaction, in this paper, we propose a Bayesian Optimization (BO) method that uses bandit human feedback in the form of complaints or satisfaction ratings to capture the quality of robot plans. However, since the trajectory space over which BO searches for a globally optimal solution can be high-dimensional, e.g., when each trajectory is parameterized by multiple waypoints, multiple queries for human feedback may be required to find a satisfactory solution, which may not be practical\cite{r9}. To reduce the dimension of the state space, the methods in \cite{kandasamy_high_2016} assume that the objective function can be decomposed into the summation of several functions that depend on variables that live in lower dimensional spaces. However, these methods assume knowledge of the objective function, which is not available in the case of human satisfaction. On the other hand, the approaches in \cite{raponi_high_2020,antonova_bayesian_2019} project the original solution space into a low dimensional space using PCA or a variational autoencoder. Since the dimension reduction in these methods does not utilize objective function information, the low dimensional space may not contain the optimal solution and, therefore, it can introduce bias. The authors in \cite{r18} combine BO with manifold Gaussian Processes (mGP) to find the optimal low dimensional space with objective function values obtained online. Nevertheless, for human-in-the-loop trajectory optimization problems with collision avoidance and dynamic feasibility constraints, this approach will still require prohibitively many queries for human feedback.

In this work, we propose a semi-supervised method to improve the data efficiency of our proposed BO method for human-in-the-loop trajectory planning. Specifically, we first utilize an autoencoder to reduce the high-dimensional space of robot trajectories to a low dimensional latent space. Since training of the autoencoder does not require human feedback, this embedding can be learnt offline in an unsupervised way. Next, we employ BO on this latent space, which we combine with human feedback collected online, to iteratively optimize both the latent space and the robot trajectories. Moreover, to improve the exploration efficiency of BO, we bias the search for new trajectories towards dynamically feasible and collision-free trajectories obtained using off-the-shelf motion planners. Our numerical experiments demonstrate that the proposed semi-supervised BO framework, combined with biased exploration, can handle diversified human preferences and achieve better data efficiency compared to existing methods.

We organize the paper as follows: 
in Section II, we formulate the proposed human-in-the-loop trajectory optimization problem. Then, we develop the proposed semi-supervised BO algorithm and present the exploration biasing technique in Section III. Simulation results are presented in Section IV. We conclude the paper in Section V.

\section{Problem Definition and Preliminaries}
\label{sec:Problem}
\subsection{Problem Definition}
Consider an environment $\mathcal{W}\subset\mathbb{R}^p$, where $p = 2 \text{ or } 3$, that contains $m$ obstacles $\mathcal{O}=\{o_1,...,o_m\}$. In addition, consider a robot with state $q\in \mathcal{Q}$, where $\mathcal{Q}$ denotes the feasible state space. Assume also that the robot satisfies the following discrete-time dynamics
\begin{align} \label{eqn:Dynamics}
q_{t +1} = f(q_{t},u_{t}),
\end{align}
where $u_t \in \mathcal{U}$ denotes the control input at time $t$ and $\mathcal{U}$ is the set of feasible control inputs. Moreover, let $x_t=\Pi(q_t)$ denotes the projection of the robot state $q_t$ to its position $x_t$ in the environment. Traditionally, the robot planning problem consists of designing a robot trajectory $\mathbf{x} = \{x_0,...,x_T\} \in \mathcal{C}\cap \mathcal{D}$ to drive the robot from a starting location $x_0$ to a goal position $x_T$, where $\mathcal{C}=\mathcal{W}\backslash\mathcal{O}$ is the space of collision-free trajectories, and $\mathcal{D}$ is the space of dynamically feasible trajectories that satisfy the dynamics in~\eqref{eqn:Dynamics}.


In this paper, we consider a robot planning problem to design robot trajectories that are not only dynamically feasible and collision-free,  but also maximize human satisfaction. Specifically, we assume that the environment $\mathcal{W}\backslash\mathcal{O}$ is populated by a group of humans in the set $\mathcal{H}$ that require different levels of assistance from the robot, which may involve adapting to specific ways that specific humans perform their tasks or avoiding actions that cause discomfort. While such requirements are subjective and hard to model mathematically in practice, the resulting levels of human satisfaction can be measured using bandit human feedback, in the form of complaints or satisfaction ratings, that expresses how desirable a robot trajectory is, without revealing the reason. To this end, we divide the humans in the set $\mathcal{H}$ into two subgroups, depending on the type of feedback they can provide. Particularly, we assume that $\mathcal{H} = \mathcal{H}^c \cup \mathcal{H}^f$, where humans $i \in \mathcal{H}^c$ provide feedback measured by the function $f_i^c: \mathbf{x}\rightarrow\{0,1\}$ that captures complaints related to discomfort so that a value of 0 indicates no discomfort, while humans $i \in \mathcal{H}^f$ provide feedback measured by the function $f_{i}^f:\mathbf{x}\rightarrow\{0,1,...,k\}$ that captures satisfaction ratings on the level of assistance provided by the robot so that lower values indicate higher levels of satisfaction. Moreover, we let the function $\mathcal{F}(\mathbf{x}) := \sum_{i=1}^{|\mathcal{H}^c|} f^c_{i}(\mathbf{x}) + \lambda\sum_{i=1}^{|\mathcal{H}^f|} f^f_{i}(\mathbf{x})$ capture the total feedback on the trajectory $\mathbf{x}$, where $\lambda$ is a weight used to ensure fairness between the two groups. Then, the problem we address in this paper is to design a collision-free and dynamically feasible trajectory $\mathbf{x}^\ast$ that maximizes human satisfaction, i.e., 
\begin{align} \label{eqn:Problem}
\mathbf{x}^\ast = \arg\min\limits_{\mathbf{x}\in \mathcal{C}\cap\mathcal{D}} \mathcal{F}(\mathbf{x}).
\end{align}
Note that the objective function $\mathcal{F}(\mathbf{x})$ is unknown. What is known is only evaluations of $\mathcal{F}(\mathbf{x})$ provided in the form of human feedback. In what follows, we employ BO to solve this planning problem. But first, we provide a brief overview of BO.

\subsection{Bayesian Optimization}

Given a dataset $D_N = [\mathbf{x}_1, \mathcal{F}(\mathbf{x}_1); \mathbf{x}_2, \mathcal{F}(\mathbf{x}_2); \dots; \mathbf{x}_N,$ $\mathcal{F}(\mathbf{x}_N)]$ consisting of $N$ sampled trajectories and corresponding bandit feedback, BO can be used to estimate the shape of the unknown objective function $\mathcal{F}$ using a non-parametric model, e.g., Gaussian Process (GP). Specifically, using GPs assumes that the function value of an unobserved trajectory $\mathbf{x}$ is subject to a Normal distribution
$\mathcal{F}(\mathbf{x}) | D_N,\mathbf{x} \sim \text{Normal}( \mu_N(\mathbf{x}), \Sigma_N(\mathbf{x})),$
where $\mu_N(\mathbf{x}) = \Sigma_0(\mathbf{x},\mathbf{x}_{1:N})
\Sigma_0(\mathbf{x}_{1:N},\mathbf{x}_{1:N})^{-1}
\mathcal{F}(\mathbf{x}_{1:N})$ is the mean, $\Sigma_N(\mathbf{x}) = \Sigma_0(\mathbf{x},\mathbf{x})
- \Sigma_0(\mathbf{x},\mathbf{x}_{1:N}) \Sigma_0(\mathbf{x}_{1:N},\mathbf{x}_{1:N})^{-1}\Sigma_0(\mathbf{x}_{1:N},\mathbf{x})$ is the variance, $\mathcal{F}(\mathbf{x}_{1:N})=[\mathcal{F}(\mathbf{x}_1),...,\mathcal{F}(\mathbf{x}_N)]$ is the collection of function evaluations, and $\Sigma_0(\mathbf{x}_{1:j},\mathbf{x}_{1:k})=
[\Sigma_0(\mathbf{x}_1,\mathbf{x}_1),...,\Sigma_0(\mathbf{x}_1,\mathbf{x}_j)
;...;
\Sigma_0(\mathbf{x}_k,\mathbf{x}_1),...,\Sigma_0(\mathbf{x}_k,\mathbf{x}_j)
]
,$ where $j,k \in \{1, 2, \dots, N\}$ is the kernel matrix. Note that, the kernel, $\Sigma_0$, is a covariance function that evaluates the similarity of two points. In this paper, we use a Gaussian kernel
$\Sigma_0(\mathbf{x},\mathbf{x}') = \sigma_f^2 \exp (-\frac{||\mathbf{x}-\mathbf{x}'||^2}{2l^2})$,
where $\sigma_f,l$ are parameters that can be tuned.
Given the estimated shape of the unknown function $\mathcal{F}(\mathbf{x})$, BO then selects the next trajectory $\mathbf{x}_e$ to evaluate so that $\mathbf{x}_e$ optimizes the acquisition function, that is, 
	$\mathbf{x}_e = \arg\min  \mu_N(\mathbf{x}) - \kappa \sqrt{\Sigma_N(\mathbf{x})}$,
where $\kappa$ is the trade-off coefficient and $\mu_N(\mathbf{x}) - \kappa \sqrt{\Sigma_N(\mathbf{x})}$ represents the lower confidential bound of the objective function value at $\mathbf{x}$.
Essentially, BO explores the trajectory $\mathbf{x}_e$ which minimizes the objective function composed of the estimated mean value $\mu_N(\mathbf{x})$ and the uncertainty of the estimation given by the variance $\Sigma_N(\mathbf{x})$ over $\mathbf{x}$.

Note that problem~\eqref{eqn:Problem} is a constrained optimization problem, where the constraint set $\mathcal{C}\cap\mathcal{D}$ cannot be represented explicitly. Therefore, it is difficult to guide the trajectory exploration procedure within the set $\mathcal{C}\cap\mathcal{D}$ using the methods proposed for constrained BO in \cite{hernandez-lobato_general_2016,gelbart_bayesian_2014,letham_constrained_2019}. Furthermore, since when the number of waypoints $T$ in the trajectory $\mathbf{x}$ is large, the dimension of the solution space $D=Tp$ is also large. Consequently, BO requires a lot of exploration and human feedback in order to find an optimal solution. In the next section, we discuss ways to address these challenges.

\section{Algorithm Design}

In this section, we develop the proposed semi-supervised BO method for human-in-the-loop robot planning. Specifically, first we reformulate problem~\eqref{eqn:Problem} so that trajectory exploration is restricted to the set of dynamically feasible and collision-free trajectories $\mathcal{C}\cap\mathcal{D}$. Then, we modify the objective function in \eqref{eqn:Problem} with additional terms that bias the search for new trajectories towards short-length, dynamically feasible, and collision-free trajectories obtained using off-the-shelf motion planners. In this way, we can improve the exploration efficiency of BO. Finally, we use an autoencoder to reduce the high-dimensional problem space into a low dimensional latent space, which we update using human feedback.

\begin{figure}[t]
\centerline{\includegraphics[width=0.75\columnwidth]{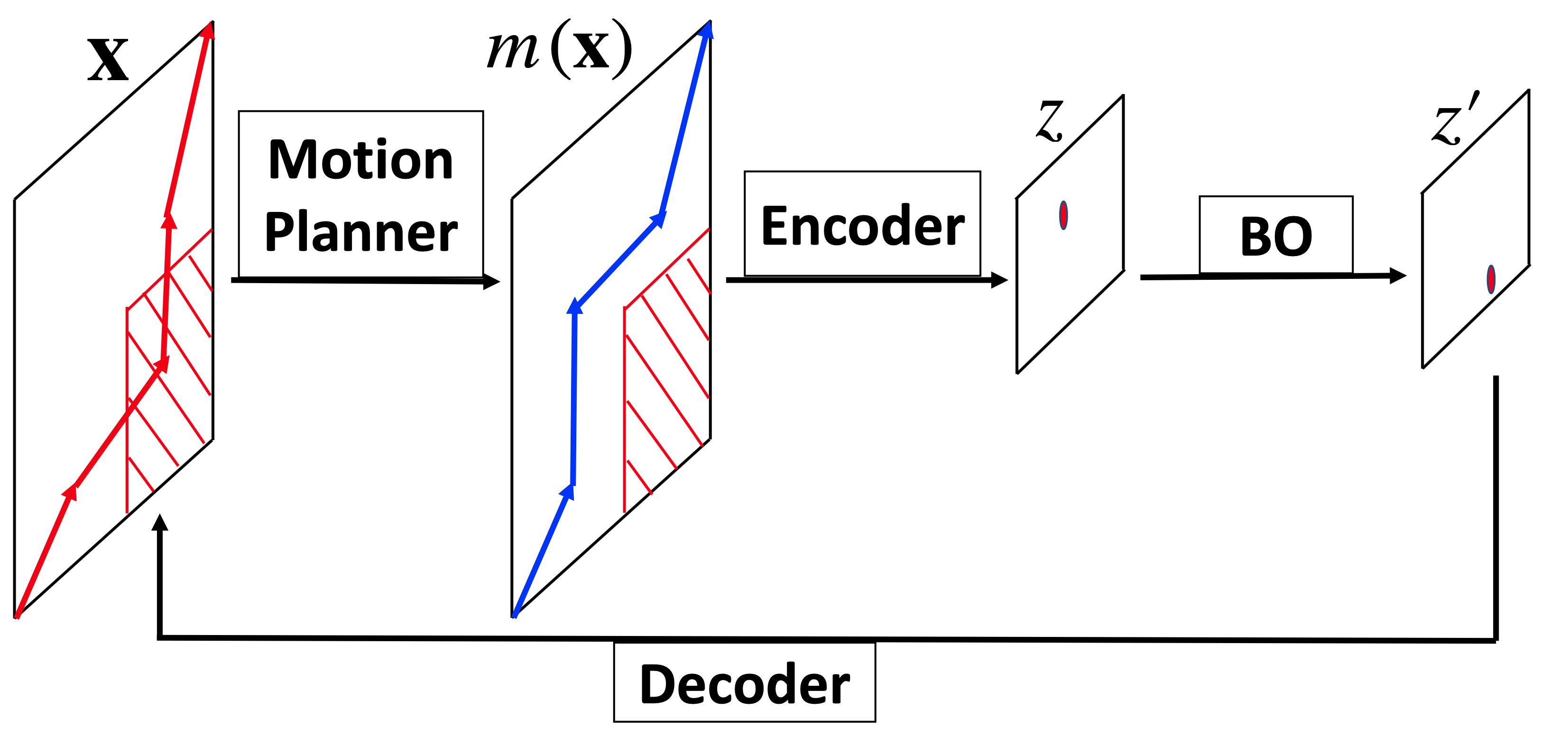}}
\caption{\small A schematic of the proposed framework. The dashed red quadrilateral represents the obstacle in 2-d space. Given a reference trajectory $\mathbf{x}$, an off-the-shelf motion planner, e.g. MPC, can be used to generate a collision-free and dynamically feasible trajectory $m(\mathbf{x})$. Then, the trajectory $m(\mathbf{x})$ is encoded into a latent space to obtain a point $z$, which is used along with human feedback to conduct GP regression and obtain the next latent point $z'$ to be evaluated. Finally, the latent variable $z'$ is decoded to generate the next reference trajectory.}
\label{fig:Pipeline}
\vspace{-4mm}
\end{figure}

\subsection{Constraint Satisfaction using Off-the-Shelf Motion Planners}

As discussed in Section II.B, restricting the exploration of new trajectories in problem (2) to the safe set $\mathcal{C}\cap \mathcal{D}$ is difficult using existing BO techniques. To this end, we reformulate problem~\eqref{eqn:Problem} as
\begin{align} \label{eqn:Reform}
	\min_{\mathbf{x} \in \mathcal{X}} \; \mathcal{F}(m(\mathbf{x})),
\end{align}
where $m(\mathbf{x}):= \mathcal{X} \rightarrow \mathcal{X}$ represents the output of an off-the-shelf motion planner, e.g., Model Predictive Control (MPC)\cite{bemporad_robust_1999,morari_model_1999,mayne_constrained_2000,camacho_model_2013} or RRT\cite{qureshi_intelligent_2015,vonasek_rrt-path_2009,jongwoo_kim_motion_2003,abbasi-yadkori_extending_2010}, and $\mathcal{X}\in\mathbb{R}^{D}$ is the space of all possible robot trajectories. Using off-the-shelf motion planners, we can map any reference trajectory $\mathbf{x}\in \mathcal{X}$ that is sampled using existing BO methods to a trajectory $m(\mathbf{x})\in \mathcal{C}\cap \mathcal{D}$ that is dynamically feasible and collision-free. 
The operation of the motion planner $m(\mathbf{x})$ on a reference trajectory $\mathbf{x}\in \mathcal{X}$ is shown in the left part of Figure~\ref{fig:Pipeline}. The reference trajectory $\mathbf{x}$ (red line) collides with the obstacle, whereas the trajectory $m(\mathbf{x})$ returned by the motion planner is collision-free. In what follows, we use MPC to obtain safe robot trajectories given reference samples $\mathbf{x}$, although other motion planners can also be used. Specifically, we solve the following MPC problem 
\begin{align} \label{eqn:MPC}
& \{\tilde{x}_{t,k}, u^*_{t,k}\} = \argmin J(\{\tilde{x}_{t,k}, u_{t,k}\}) \nonumber \\
& \text{s.t. } q_{t,k+1} = f(q_{t,k},u_{t,k}), \; u_{t,k} \in \mathcal{U}, \\
& \quad \;\; \Pi(q_{t,0}) = x_t', \; \Pi(q_{t,k}) \in \mathcal{C}, \text{ for } k = 1, \dots, K-1, \nonumber
\end{align}
where $K$ is the control horizon, $\tilde{x}_{t,k} = \Pi(q_{t,k})$, is the projection of the current state of the robot to the space of robot positions, $J(\{\tilde{x}_{t,k}, u_{t,k}\}) := \sum_{k=0}^{K} (\tilde{x}_{t,k}- x_{t+k})^T Q_k (\tilde{x}_{t,k}- x_{t+k}) + \sum_{k=0}^{K-1} u_{t,k}^T R_k u_{t,k}$ represents the accumulated tracking error of the reference trajectory $\mathbf{x}$ and control energy cost over the future $K$ steps, and $x_t'$ denotes the true position of the robot at time step $t$. Note that since the dynamic feasibility and obstacle avoidance constraints are explicitly encoded in \eqref{eqn:MPC}, the output of the motion planner $m(\mathbf{x})$ is guaranteed to belong to the set $\mathcal{C}\cap\mathcal{D}$. Other safety requirements can also be similarly encoded into problem~\eqref{eqn:MPC}.


\begin{figure}[t] 
\centerline{\includegraphics[width=0.4\columnwidth]{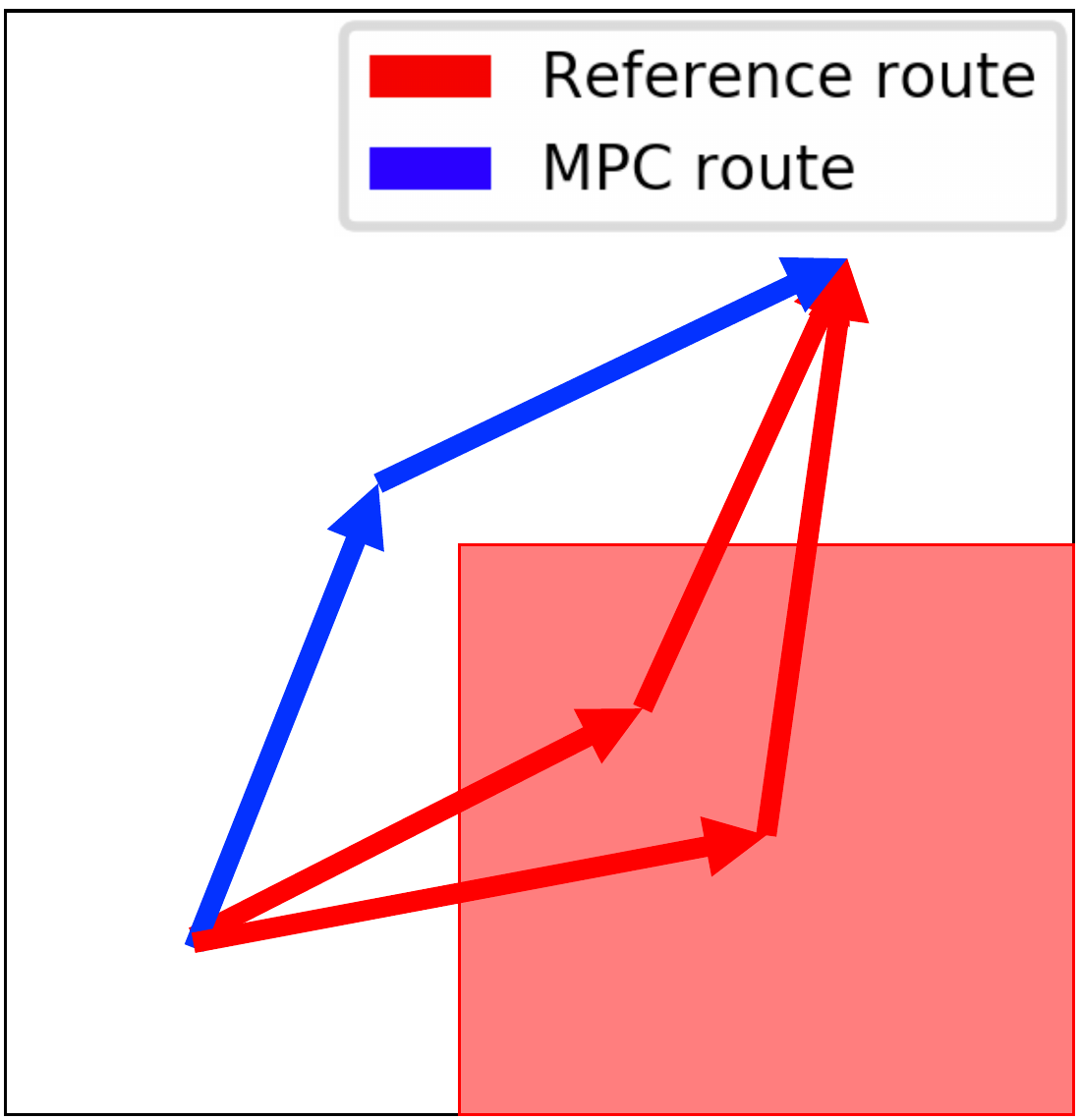}}
\caption{\small An example where two different reference trajectories (red lines) are mapped to the same trajectory (blue line) using MPC controller~\eqref{eqn:MPC}. The red rectangle represents the obstacle.}
\label{fig:MPC}
\vspace{-4mm}
\end{figure}

\subsection{Exploration Bias in Bayesian Optimization}
Notice that the use of motion planners to map unsafe samples $\mathbf{x}\in \mathcal{X}$ to safe trajectories $m(\mathbf{x})$ may reduce the sampling efficiency of BO, since it is possible that multiple unsafe samples $\mathbf{x}$ and $\mathbf{x}'$ from the set $\mathcal{X}$ are mapped to the same safe trajectories, i.e., $m(\mathbf{x}) = m(\mathbf{x}')$. An example is shown in Figure ~\ref{fig:MPC}. To improve the sample efficiency of BO to solve problem \eqref{eqn:Reform}, in this section, we propose a way to bias the exploration of new trajectories $\mathbf{x}\in \mathcal{X}$ towards dynamically feasible and collision-free trajectories that are also feasible solutions of typical motion planners. This way, we can avoid sampling redundant unsafe trajectories $\mathbf{x}\in \mathcal{X}$ that are mapped to the same safe trajectory $m(\mathbf{x})$.

To bias trajectory exploration in problem~\eqref{eqn:Reform}, we introduce the term $\|\mathbf{x}-m(\mathbf{x})\|$ in the objective function which allows BO to learn the subspace in $\mathcal{X}$ that contains trajectories that can be tracked by the motion planner~\eqref{eqn:MPC}. Effectively, by selecting reference trajectories $\mathbf{x}$ that minimize the term $\|\mathbf{x}-m(\mathbf{x})\|$, we obtain that, different reference trajectories $\mathbf{x}$ and $\mathbf{x'}$ in this subspace will be mapped to different trajectories $m(\mathbf{x})$ and $m(\mathbf{x}')$ since both $\mathbf{x}$ and $\mathbf{x}'$ are feasible solutions of the motion planner. We note that the term $\|\mathbf{x}-m(\mathbf{x})\|$ does not affect the solution of the original problem~\eqref{eqn:Reform}. To see this, note that the global optimizer $\mathbf{x}^\ast = \argmin_{\mathbf{x} \in \mathcal{X}} \; \mathcal{F}(m(\mathbf{x}))$ also minimizes the modified problem $\min_{\mathbf{x} \in \mathcal{X}} \; \mathcal{F}(m(\mathbf{x})) + \rho\|\mathbf{x}-m(\mathbf{x})\|$, where $\rho$ is a penalty parameter. This is because, if the planning problem is feasible, then there exists a minimizer $\mathbf{x}^\ast \in \mathcal{X}$ that can be perfectly tracked by the motion planner. 


In practice, it may also be useful to add additional terms to the objective function in problem~\eqref{eqn:Reform}
that specify soft constraints on the robot's trajectories. For example, we can modify the objective function in \eqref{eqn:Reform} by adding the term $l(m(\mathbf{x}))$ denoting the total length of the trajectory $m(\mathbf{x})$. This modification will bias BO exploration towards trajectories with shorter length. However, unlike the tracking error $\|\mathbf{x}-m(\mathbf{x})\|$, the inclusion of the term $l(m(\mathbf{x}))$ in the objective of the problem \eqref{eqn:Reform} affects the final solution, because the global optimizer $\mathbf{x}^\ast = \argmin_{\mathbf{x} \in \mathcal{X}} \; \mathcal{F}(m(\mathbf{x}))$ does not necessarily minimize the modified problem $\min_{\mathbf{x} \in \mathcal{X}} \; \mathcal{F}(m(\mathbf{x})) + \tau l(m(\mathbf{x}))$, where $\tau$ is a penalty parameter. Therefore, the parameter $\tau$ needs to be carefully chosen that human satisfaction is not significantly affected.
In what follows, we reformulate problem (6) to the problem
\begin{align} \label{eqn:Reform_Augment}
\mathbf{x}^*=\arg\min \limits_{\mathbf{x}}  Obj(\mathbf{x}),
\end{align}
where $ Obj(\mathbf{x})=\mathcal{F}(m({\mathbf{x}})) + \rho||\mathbf{x}-m(\mathbf{x})|| + \tau l(m(\mathbf{x}))$ so that exploration in BO can be accelerated towards dynamically feasible, collision-free, and short-length trajectories. 

\begin{algorithm}[t]
	\caption{Online Semi-supervised Bayesian Optimization}
	\label{alg:SemiSuper}
	\begin{algorithmic}[1]
		\REQUIRE ~~\\
		Black box objective function $f:\mathbb{R}^D \rightarrow \mathbb{R}$, unlabeled data set $\mathcal{D}_{un} \in \mathbb{R}^{N_u \times D}$,
		embedding dimension $d$, maximum query number $N_{Query}$ and size of mini-batch $N_m$.\\
		\STATE{Randomly initialize $\theta_{E}$, $\theta_{D}$, $\theta_{G}$;}
		\STATE{Train autoencoder with unlabeled data and update $\theta_{E}$, $\theta_{D}$;}
		\STATE{Randomly generate an initial reference trajectory $\mathbf{x}_0$}
		\STATE{Use motion planner to get $m(\mathbf{x}_0)$ and evaluate human feedback to get $\mathcal{D}_l=\{\{m(\mathbf{x}_0),y_0\}\}$}
		\FOR{$1\leq i\leq N_{Query}$}
		\STATE{Encode $\mathcal{D}_l$ to $\mathcal{D}^e_l$}
		\STATE{Conduct GP regression with $\theta_G$ on $\mathcal{D}^e_l$ to get acquisition function $q:\mathbb{R}^d \rightarrow \mathbb{R}$}
		\STATE{${z}_{i} = \argmax\limits_{z\in \mathbb{R}^{d}} \; q(z)$}
		\STATE{Let $\mathbf{x}_i = D(z_i)$}
		\STATE{Use motion planner to get $m(\mathbf{x}_i)$ and evaluate $y_{i} = Obj(m(\mathbf{x}_i))$}
		\STATE{$\mathcal{D}_l=\mathcal{D}_l \cup \{\{ m(\mathbf{x}_i)),y_{i}\}\}$}
		\IF{$i\%N_m=0$}
		\STATE{Update $\theta_{E}$, $\theta_{D}$ and $\theta_{G}$ according to Algorithm 2;
		}
		\ENDIF
		\ENDFOR
		\STATE{Output $\mathbf{x}_{N_{Query}}$.}
	\end{algorithmic}
\end{algorithm}

\subsection{Semi-Supervised Dimension Reduction and Trajectory Optimization}
Problem~\eqref{eqn:Reform_Augment} improves the exploration efficiency of BO within the constraint set $\mathcal{C}\cap\mathcal{D}$. However, the safe set $\mathcal{C}\cap\mathcal{D}$ is of the same high dimension as the original space $\mathcal{X}$. 
In this section, we first discuss how to reduce the dimension of the original space $\mathcal{X}$ by embedding it into a lower dimensional latent space $\mathcal{Z}$, which contains regular trajectories that are of certain degree of smoothness. This embedding is trained using trajectory data not labeled by human feedback. Therefore, this step can be completed offline and in an unsupervised way. Then, we run BO on the low dimensional latent space. Specifically, we sample new trajectories in an online way, query humans for feedback on each new trajectory, and use these data to update both the latent space and the GP model in an online way. Since this process uses human feedback, it is supervised in nature. The complete semi-supervised trajectory optimization framework is presented in Algorithm~\ref{alg:SemiSuper}. Next, we explain each step in details. 


Given a pair of starting and goal robot positions, we can use off-the-shelf motion planners, e.g., RRT, to generate a dataset $\mathcal{D}_{un}$ containing a large number of random trajectories that respect the robot dynamics~\eqref{eqn:Dynamics}\footnote{We note that the generated latent space can be used for any pair of starting and goal positions. To see this, we can always scale, translate and rotate the trajectory decoded from the latent space appropriately, so that the starting and goal positions are as desired. If we also know the position of obstacles when training the autoencoder, we can use a motion planner to construct a dataset $\mathcal{D}_{un}$ that contains trajectories which are both dynamically feasible and collision free.}. Then, we use the autoencoder developed in \cite{r25} to embed these randomly generated trajectories to a low dimensional latent space. Specifically, consider an encoder function $E_{\theta_E} := \mathcal{X} \rightarrow \mathcal{Z}$, where $\theta_E$ is the parameter of the encoder function and $\mathcal{Z} \in \mathbb{R}^d$ is the latent variable space of dimension $d$, where $d \ll D$. In addition, consider a decoder function $D_{\theta_D}:= \mathcal{Z} \rightarrow \mathcal{X}$, where $\theta_D$ is the decoder parameter. Then, we can train the encoder and decoder models by solving the problem
\begin{align} \label{eqn:Reconstruct}
\min_{\theta_D, \theta_E}   L_{re}(\theta_D, \theta_E; \mathcal{D}_{un}),
\end{align}
where $L_{re}(\theta_D, \theta_E; \mathcal{D}_{un}) := \sum_{ \mathbf{x} \in \mathcal{D}_{un}} ||\mathbf{x} -D_{\theta_D}(E_{\theta_E}(\mathbf{x}))||_2^2$ measures the reconstruction loss between the original trajectory $\mathbf{x}$ and the output of the autoencoder $D_{\theta_D}(E_{\theta_E}(\mathbf{x}))$ (line 2 in Algorithm~\ref{alg:SemiSuper}). Since training the autoencoder does not require human feedback on the trajectories in the set $\mathcal{D}_{un}$, this process is unsupervised.

Next, we run BO on the learnt latent space $\mathcal{Z}$ . Specifically, instead of directly constructing a GP model of the unknown objective function $Obj$ in \eqref{eqn:Reform_Augment} that depends on the trajectory $\mathbf{x}$, we construct a GP model that depends on the latent variable $z$. 
As shown in Figure~\ref{fig:Pipeline}, given a newly sampled reference trajectory $\mathbf{x}$ (lines 3 and 9 in Algorithm~\ref{alg:SemiSuper}), we first use the motion planner~\eqref{eqn:MPC} to generate a safe robot trajectory $m(\mathbf{x})$ and collect the human feedback $\mathcal{F}(m(\mathbf{x}))$ (lines 4 and 10 in Algorithm~\ref{alg:SemiSuper}). Then we encode the trajectory $m(\mathbf{x})$ into a point $z$ in the latent space (line 6 in Algorithm~\ref{alg:SemiSuper}). With the latent point $z$ and the human feedback, we conduct GP regression to get the next latent point $z'$ to be evaluated (lines 7 and 8 in Algorithm~\ref{alg:SemiSuper}). Subsequently, we decode $z'$ into the original trajectory space $\mathcal{X}$ to get a new reference trajectory (line 9 in Algorithm~\ref{alg:SemiSuper}) and the process repeats.


\begin{algorithm}[t]
	\caption{Parameter Update}
	\label{alg:Param}
	\begin{algorithmic}[1]
		\REQUIRE ~~\\
		Initial parameter value $\theta_{E}$, $\theta_{D}$, $\theta_{G}$, labeled data set $\mathcal{D}_l=\{\cup_{i=1}^{N} \{ \mathbf{x}_i, y_i\}\} $, number of epoch $N_e$, and step size $\alpha_E,\alpha_D$ and $\alpha_G$;\\
		\FOR{$1 \le k \le N_e$}
		\STATE{$\theta_{E} = \theta_{E} - \alpha_E\nabla_{\theta_{E}}(L_{re} + L_{nmll})$}
		\STATE{$\theta_{D} = \theta_{D} - \alpha_D\nabla_{\theta_{D}}L_{re}$ }
		\STATE{$\theta_{G} =\theta_{G} - \alpha_G\nabla_{\theta_{G}}L_{nmll}$}
		\ENDFOR
		\STATE{Output $\theta_E$, $\theta_D$ and $\theta_G$.}
	\end{algorithmic}
\end{algorithm}

Note that since the initial latent space $\mathcal{Z}$ is learnt offline in an unsupervised way without human feedback, it does not necessarily contain the true optimal solution to problem~\eqref{eqn:Reform_Augment}. Therefore, as soon as $N_m$ new data samples have been collected using BO, they are used to update the parameters of the autoencoder, $\theta_E$ and $\theta_D$, and the GP kernel parameter $\theta_G$, i.e., $[\sigma_f,l]$ in Section II.B. Specifically, the GP parameter $\theta_G$ is updated by minimizing the
negative marginal log-likelihood function
\begin{align}{\label{l_nmll}}
& L_{nmll} = -\text{log } p(Y|\mathbf{X},\theta_G,\theta_E) \\
&=Y^T\Sigma_0^{-1}(\mathbf{x}_{1:N_m},\mathbf{x}_{1:N_m})Y + \frac{1}{2}\log |\Sigma_0(\mathbf{x}_{1:N_m},\mathbf{x}_{1:N_m})|, \nonumber
\end{align}
where $Y=[Obj(\mathbf{x}_1),...,Obj(\mathbf{x}_{N_m})]^T$ and $X=[\mathbf{x}_1,...,$ $\mathbf{x}_{N_m}]$.
Moreover, the autoencoder parameters $\theta_E, \theta_D$ are updated so that the reconstruction loss in \eqref{eqn:Reconstruct} is minimized with respect to the new data in $\mathcal{D}_{N_m}$. 
We run $N_e$ iterations of stochastic gradient decent to update these parameters according to Algorithm~\ref{alg:Param}, and then run BO again to collect new data samples with the new autoencoder and GP parameters.
Since the updates in Algorithm \ref{alg:Param} are conducted using trajectory data that contain human feedback, this procedure is supervised. 

Combining the unsupervised and supervised learning process described above, we obtain the semi-supervised trajectory optimization framework presented in Algorithm~\ref{alg:SemiSuper}. 

\section{Numerical Experiments}
In this section, we corroborate the effectiveness of the proposed semi-supervised trajectory optimization framework on the human-in-the-loop robot planning example discussed in Section~\ref{sec:Problem}. Moreover, we demonstrate the merit of exploration bias in Section III.B and semi-supervised learning in Section III.C through ablation studies. All the numerical experiments in this section are conducted using Python 3.63 on a computer with 2.6 GHz Intel Core i7 processor and 16GB RAM.

Consider a workspace $\mathcal{W} \in \mathbb{R}^2$ of size $20\times20$. Two square obstacles of length $2$ are located at locations $(5, 8)$ and $(11,13)$ in the workspace. The goal is to drive a robot from position $(0,0)$ to its goal position $(20,20)$.
At time $t$, the robot state is denoted by $q_t = [p_{x,t}, p_{y,t}, \theta_t]^T$, where $q_{x,t}$ and $q_{y,t}$ are the robot's coordinates in the workspace and $\theta_t$ is the orientation of the robot. The control signal of the robot at time $t$ is denoted by $u_t = [v_t, \omega_t]^T$, where $v_t$ represents the linear velocity and $\omega_t$ denotes the angular velocity.
The dynamics of the robot are subject to the discrete-time unicycle model \cite{luo_socially-aware_2020}, i.e.,
\begin{equation} \label{eqn:Unicycle}
q_{t+1} = q_{t} +  \Delta t 
\begin{bmatrix}
\cos\theta_{t} & 0\\
\sin\theta_{t} & 0\\
0 & 1
\end{bmatrix}
u_{t},
\end{equation}
where $\Delta t$ is the time interval.
%
%
  Consider also a group of humans in the set $\mathcal{H}=\mathcal{H}^c \cup \mathcal{H}^f$ randomly distributed in the collision-free workspace $\mathcal{C}$. Let
$|\mathcal{H}^c| = 5$ and $|\mathcal{H}^f| = 20$. We assume that for every human $i \in \mathcal{H}^f$, the feedback function $f_i^f(\mathbf{x}) = 1$ if the robot trajectory passes through a ball of radius $r_f$ centered at that human's position $p_i$. Otherwise, $f_i^f(\mathbf{x}) = 0$. Essentially, humans in the set $\mathcal{H}^f$ complain if the robot trajectory comes too close to them. On the other hand, if $i \in \mathcal{H}^c$, then $f_i^c(\mathbf{x})= \min\{\text{floor}(\text{dist}(\mathbf{x},p_i),r_c) , k\}$, where $\text{dist}(\mathbf{x}, p_i)$ is the distance from $p_i$ to the trajectory $\mathbf{x}$ and we select $k=5$. This feedback function models a $(k+1)$-level rating on how satisfied a human is about robot's trajectory. If the robot is within distance $r_c$ from the human, a rate $0$ is given. On the contrary, if the robot trajectory is far from human, a rate $5$ is sent. Essentially, humans in $\mathcal{H}^c$ want the robot to come close. Note that no contextual information on the feedback functions $f_i^f$ and $f_i^c$, e.g., the human positions $p_i$ or ranges $r_f$ and $r_c$, is assumed to be known by our semi-supervised BO method. All that is known is the feedback functions' evaluations.

We used a single layer fully connected neural network with sigmoid activation functions as our Encoder and Decoder. The unsupervised training set $\mathcal{D}_{un}$ was generated using RRT\cite{qureshi_intelligent_2015,vonasek_rrt-path_2009,jongwoo_kim_motion_2003,abbasi-yadkori_extending_2010} so that the routes are in the collision-free and dynamically-feasible space $\mathcal{C}\cap\mathcal{D}$. Since RRT does not involve the optimization of any utility, it generates trajectories that span the entire domain when sufficiently many samples are collected.
\begin{figure}[t] 
	\centerline{\includegraphics[width=0.65\columnwidth]{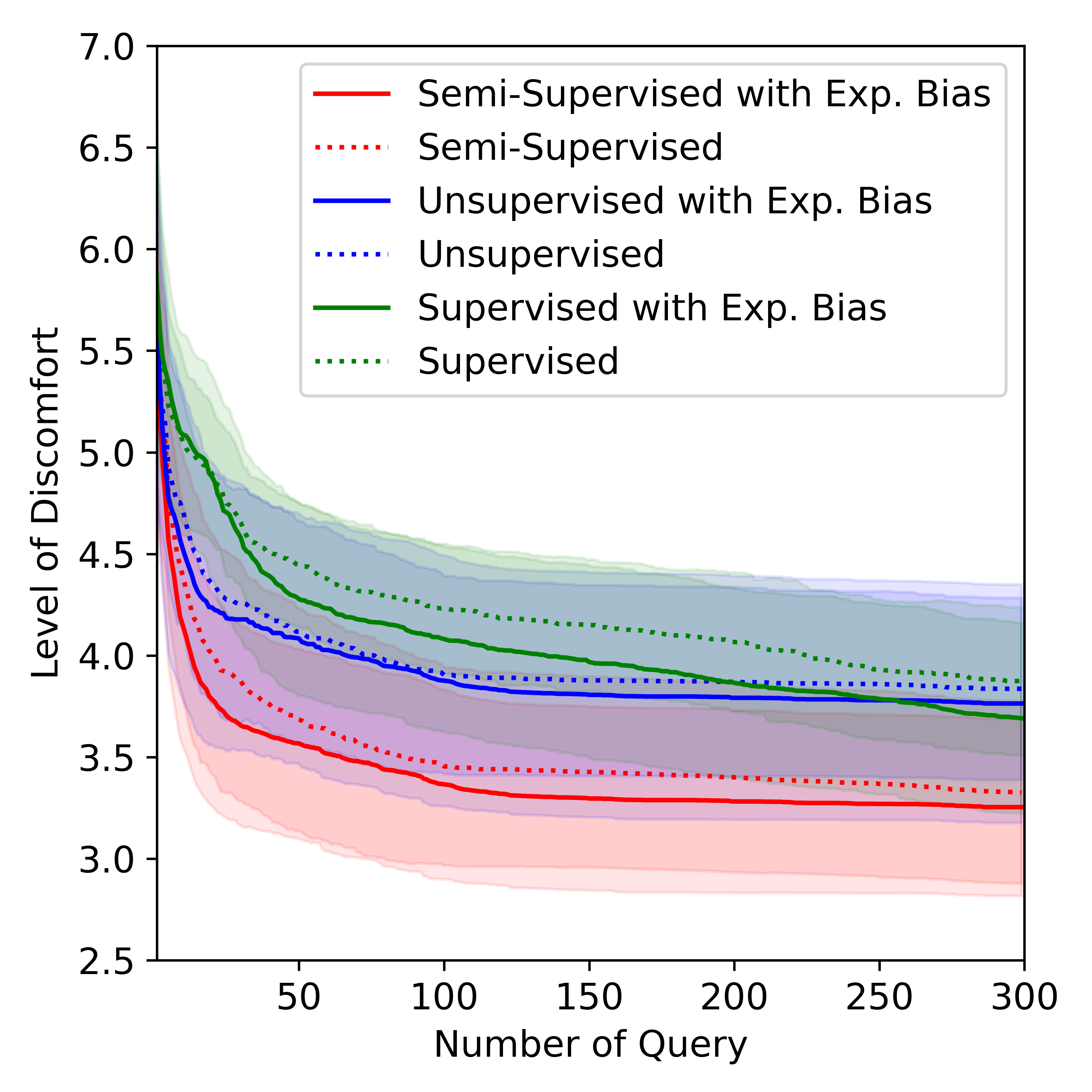}}
	\caption{\small Comparison of the human satisfaction $\mathcal{F}(m(\mathbf{x}))$ returned by supervised learning (green curves),  unsupervised learning (blue curves), and semi-supervised learning (red curves). Solid curves represent implementation of these algorithms with exploration bias, while dashed curves are without exploration bias. The horizontal axis represents the number of queries for human feedback.}
	\label{fig:Comparison}
\vspace{-4mm}	
\end{figure}
In what follows, we compare our semi-supervised BO method in Algorithm \ref{alg:SemiSuper} to a \textit{supervised learning} approach where the trajectory is optimized using only the online data associated with human feedback, without training the autoencoder using the unlabeled dataset. We also compare our method to an \textit{unsupervised learning} approach where the autoencoder is trained using the unlabeled dataset beforehand, and BO is run on the fixed latent space given by the trained autoencoder. In our simulations, we parametrize the robot trajectories by $20$ waypoints and, therefore, the dimension of the original planning problem is $40$.

In Figure~\ref{fig:Comparison} we study the effect of our proposed biased exploration method. Specifically, we compare the performance of supervised, unsupervised and semi-supervised learning with and without the exploration bias $\|\mathbf{x} - m(\mathbf{x})\|$.
We observe that exploration bias improves the performance of all three learning methods compared to their counterparts without bias.  
The benefit of biased search in BO is more apparent in the case of supervised learning. This is because the latent space in the case of supervised learning is randomly initialized and, therefore, it is decoded into many trajectories that have collisions and are dynamically infeasible. In comparison, the latent spaces in the case of unsupervised and semi-supervised learning are trained using collision-free and dynamically feasible trajectories.
Furthermore, we observe in Figure~\ref{fig:Comparison} that semi-supervised learning (red curves) outperforms both supervised learning (green curves) and unsupervised learning (blue curves) with or without the exploration bias. Specifically, unsupervised learning gets trapped at suboptimal trajectories and is finally outperformed by the supervised approach. This is because unsupervised learning is initialized with a latent space trained using randomly generated collision-free and dynamically feasible trajectories, which accelerates the trajectory search in the beginning.
However, since this latent space is not trained using all possible trajectories that satisfy the constraints and is not updated using online data, the final solution contains bias.
On the other hand, the proposed semi-supervised learning method finds the globally optimal trajectory faster than the other two methods.
The reason is that semi-supervised learning uses the same initial latent space as that used by the unsupervised method to achieve efficient exploration in the beginning, but it also updates this latent space using human feedback to remove the bias that is present in the unsupervised way. Note also that semi-supervised learning achieves better performance compared to the supervised and unsupervised methods using less than 30 queries for human feedback. This number is comparable to our recent zeroth-order method~\cite{luo_socially-aware_2020} which, however, can not handle different types of human feedback and is prone to local minima, and much fewer than queries needed by typical RL methods like those discussed in Section I. Therefore, our approach is also feasible in practice.

\begin{figure}[t]
	\centering
	\includegraphics[width=0.55\columnwidth]{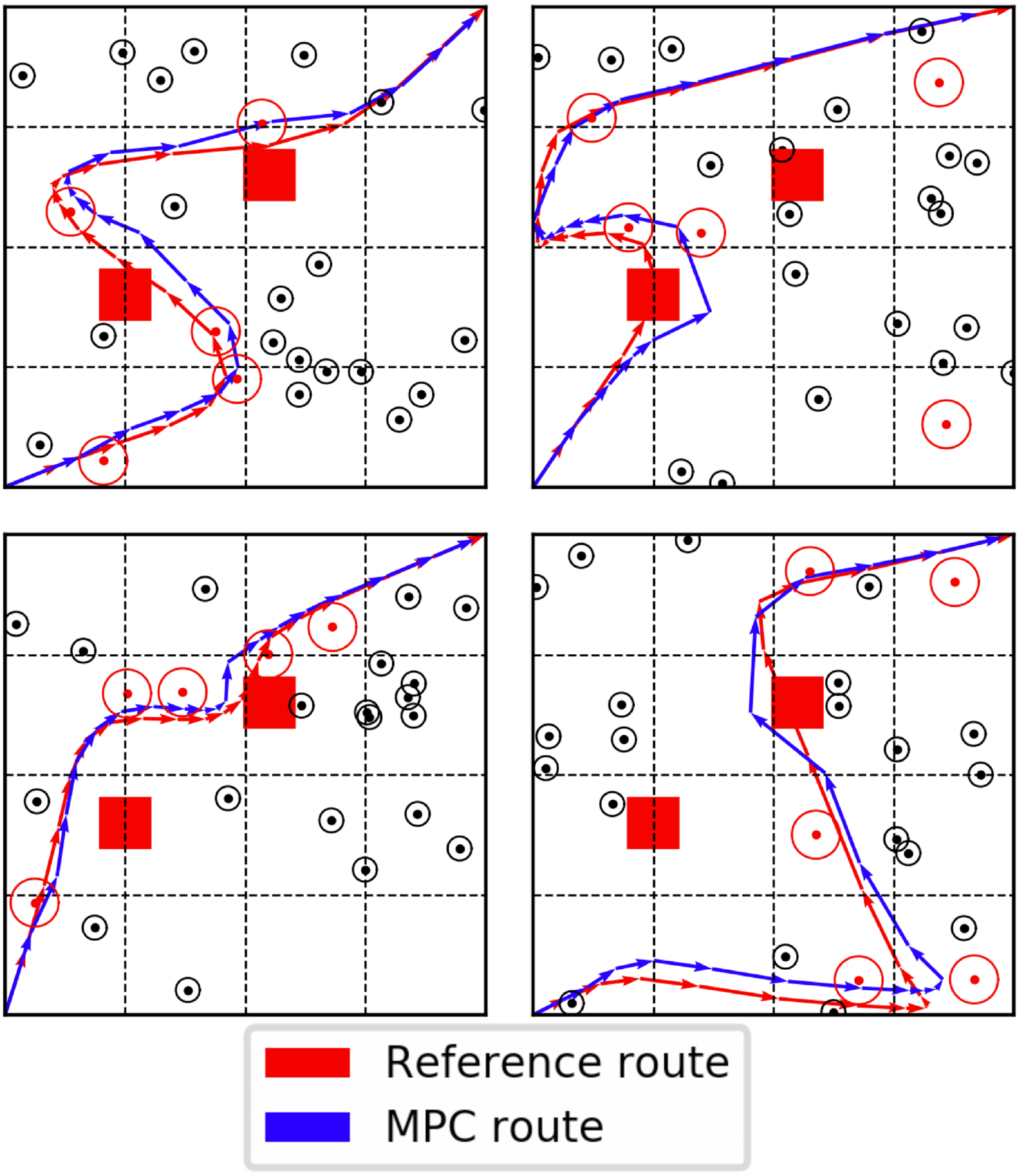}
	\caption{\small Plots of the reference trajectories $\mathbf{x}$ (red line) and the actual robot trajectories $m(\mathbf{x})$ (blue line) returned by the MPC controller~\eqref{eqn:MPC} for four different configurations of the humans in the workspace. All trajectories are obtained after 50 queries for human feedback  using semi-supervised learning with exploration bias. The red circles represent humans in group $\mathcal{H}^c$ and the black circles represent humans in group $\mathcal{H}^f$.}
	\label{fig:trajectory}
\vspace{-4mm}
\end{figure}

In Figure~\ref{fig:trajectory}, we present the reference trajectories $\mathbf{x}$ and the actual trajectories $m(\mathbf{x})$ followed by the robot for four different configurations of the humans $\mathcal{H}^c \cup \mathcal{H}^f$ in the workspace. 
These trajectories are returned by our semi-supervised learning method using 50 rounds of human feedback. These trajectories satisfy the requirements of most humans except for very few cases due to the use of the trajectory length term in \eqref{eqn:Reform_Augment}. As discussed in Section III.B, the objective in problem~\eqref{eqn:Reform_Augment} strikes a balance between human satisfaction and trajectory length, which causes remaining complaints or poor ratings.

\section{Conclusion}
In this paper, we proposed a semi-supervised Bayesian Optimization (BO) method for human-in-the-loop motion planning that relies only on bandit human feedback, in the form of complaints or satisfaction ratings, that express how desirable a trajectory is without revealing the reason. To reduce the number of queries for human feedback needed by BO to solve this high-dimensional trajectory optimization problem, we used an autoencoder to map the high-dimensional trajectory space to a low dimensional latent space, which we updated using human feedback. Moreover, to improve the exploration efficiency of BO, we proposed a way to bias the exploration for new trajectories in BO towards dynamically feasible and collision-free trajectories that satisfy the problem constraints. 
We demonstrated the efficiency of our proposed trajectory planning method in a scenario with humans that have diversified and unknown demands and showed that it can find better solutions than competitive methods with very few queries for human feedback.

\newpage
\bibliographystyle{IEEEtran}
\bibliography{refs}

\vspace{12pt}
\color{red}

\end{document}